\documentclass[runningheads]{llncs}

\usepackage[year=2026]{eccv}

\usepackage{eccvabbrv}

\usepackage{graphicx}
\usepackage{booktabs}
\usepackage{multirow}
\usepackage[shortlabels]{enumitem}
\usepackage{placeins}
\usepackage{float}

\usepackage[accsupp]{axessibility}  

\usepackage{hyperref}

\newcommand{\highlightr}[1]{%
  {\setlength{\fboxsep}{1pt}%
  \colorbox[HTML]{F4CCCC}{\strut #1}}%
}

\newcommand{\highlighto}[1]{%
  {\setlength{\fboxsep}{1pt}%
  \colorbox[HTML]{FCE5CD}{\strut #1}}%
}

\newcommand{\highlighty}[1]{%
  {\setlength{\fboxsep}{1pt}%
  \colorbox[HTML]{FFF2CC}{\strut #1}}%
}

\begin{document}

\title{MAFNet-UR\textsuperscript{3}: Mask-Aware Fusion Network for Unified Raindrop and Reflection Removal}

\titlerunning{MAFNet-UR\textsuperscript{3} for Unified Raindrop and Reflection Removal}

\providecommand{\iflatexml}{\iffalse}

\iflatexml

\author{Yeongjin Jeong\inst{1} \and
Hyunjin Cho\inst{2} \and
Minsang Kim\inst{3} \and
Jiwon Lee\inst{4} \and
Jae Woong Soh\inst{5}}

\institute{
Department of AI,
Gwangju Institute of Science and Technology (GIST),
Gwangju, Republic of Korea\\
\email{jyjin1206@gm.gist.ac.kr}
\and
Department of EECS,
Gwangju Institute of Science and Technology (GIST),
Gwangju, Republic of Korea\\
\email{roycecho@gm.gist.ac.kr}
\and
Department of EECS,
Gwangju Institute of Science and Technology (GIST),
Gwangju, Republic of Korea\\
\email{minsangkimug@gm.gist.ac.kr}
\and
Department of AI,
Gwangju Institute of Science and Technology (GIST),
Gwangju, Republic of Korea\\
\email{wwwljw2@gm.gist.ac.kr}
\and
Department of AI and Department of EECS,
Gwangju Institute of Science and Technology (GIST),
Gwangju, Republic of Korea\\
\email{jaewoongsoh@gist.ac.kr}\\
Corresponding author.
}

\else

\author{Yeongjin Jeong\inst{1} \and
Hyunjin Cho\inst{2} \and
Minsang Kim\inst{2} \and
Jiwon Lee\inst{1} \and\\
Jae Woong Soh\inst{1,2}\thanks{Corresponding author.}}

\institute{\textsuperscript{1}Department of AI;
\textsuperscript{2}Department of EECS\\
Gwangju Institute of Science and Technology (GIST), Gwangju, Republic of Korea\\
\email{\{jyjin1206, roycecho, minsangkimug, wwwljw2\}@gm.gist.ac.kr,
jaewoongsoh@gist.ac.kr}}

\fi

\authorrunning{Y.~Jeong et al.}

\maketitle

\begin{abstract}
Raindrops and reflections frequently co-occur on windshields and jointly obscure scene content, making their unified removal, referred to as UR\textsuperscript{3}, a highly ill-posed problem.
Existing approaches are mainly based on diffusion models and thus can produce visually plausible restorations. However, they may generate scene-inconsistent structures in severely corrupted regions.
To address this limitation, we propose MAFNet-UR\textsuperscript{3}, which adopts a three-stage framework to exploit multiple observations for scene-consistent unified raindrop and reflection removal. Specifically, Stage I suppresses raindrops while preserving the reflection layer, and Stage II performs prior-guided reflection removal followed by image-space refinement. Stage III estimates degradation masks from intermediate restoration results to provide degradation cues and performs mask-aware multi-scale fusion across observations to preserve scene structure and fine details.
As a result, our proposed MAFNet-UR\textsuperscript{3}-L achieves state-of-the-art performance on the RDRF-testing benchmark, outperforming DiffUR\textsuperscript{3} by 2.33 dB in PSNR and 2.7634 in the challenge score. Our method ranked \textbf{third} in the LoViF 2026 UR\textsuperscript{3} Challenge, achieving a final score of 41.1416.
\end{abstract}

\section{Introduction}
\label{sec:introduction}

Raindrops and reflections often co-occur on windshields in rainy environments, degrading images captured by on-board cameras. Raindrops usually introduce local blur and occlusion, while reflections produce semi-transparent structures that overlap with the transmission. These degradations can obstruct object recognition in autonomous driving scenarios, posing a significant challenge for reliable visual perception. The Unified Removal of Raindrops and Reflections (UR\textsuperscript{3}) task aims to restore clean scenes from images degraded by both raindrops and reflections.

Recently, DiffUR\textsuperscript{3}~\cite{liu2026diffur3} addresses the UR\textsuperscript{3} task, using a two-stage restoration pipeline that leverages the generative power of diffusion models~\cite{ho2020denoising, rombach2022high}.
However, diffusion-based restoration may hallucinate realistic content that is inconsistent with the actual scene in severely corrupted regions, which reduces pixel-level fidelity.

To tackle the UR\textsuperscript{3} task, we propose MAFNet-UR\textsuperscript{3}, a multi-stage, multi-observation restoration framework designed to prioritize scene-consistent reconstruction over unconstrained content generation.
Empirically, we find that reflection removal is substantially more challenging than raindrop removal, as reflection layers often contain semantically meaningful structures that are difficult to disentangle from the transmitted scene.
This ambiguity makes the unified removal problem highly ill-posed. We therefore decompose the task and exploit multiple observations of the same scene, leveraging the transient nature of raindrops to obtain complementary realizations. By combining this information, our approach suppresses degradations while preserving the fidelity of the underlying scene.

Specifically, the proposed framework consists of multiple stages.
First, for each input observation, it removes raindrops while retaining the reflection layer for the subsequent stage, and then performs prior-guided reflection removal.
Second, degradation masks are estimated from the differences between consecutive images in the restoration cascade.
The degraded observations, cascade-restored images, and estimated masks are then used to predict spatial confidence maps and perform weighted multi-scale feature fusion across observations. Finally, a weighted combination of the restored images forms a restoration anchor, which is refined using a magnitude-constrained residual to recover corrupted regions while preserving the underlying scene structure. Together, these components form a multi-observation framework that combines mask-aware spatial confidence estimation with anchor-based residual refinement. We present two variants: MAFNet-UR\textsuperscript{3}-B and MAFNet-UR\textsuperscript{3}-L, denoting the Base and Large models, respectively.

The experimental results demonstrate that our \textbf{MAFNet-UR\textsuperscript{3}} achieves state-of-the-art performance on the RDRF-testing benchmark, outperforming DiffUR\textsuperscript{3}~\cite{liu2026diffur3} across various metrics. Additionally, qualitative comparisons show that the proposed method effectively reduces scene-inconsistent content generated by diffusion priors and produces reconstructions that better preserve the underlying scene. Furthermore, our method achieved a challenge score of 41.1416 and ranked third on the LoViF Challenge test set.

\section{Related Work}
\label{sec:related-work}

\subsection{Raindrop Removal}
\label{sec:related-raindrop-removal}

Early deep learning approaches to raindrop removal relied primarily on convolutional networks and were later extended to handle more complex rain degradations~\cite{eigen2013restoring, quan2019deep,quan2021removing}. In parallel, a GAN-based method combined a raindrop-focused attention mechanism with adversarial training to improve the perceptual naturalness of restored images~\cite{qian2018attentive}. Transformer-based methods later improved restoration performance by modeling long-range degradation relationships~\cite{chen2023sparse}. Beyond raindrop-specific methods, all-in-one adverse-weather restoration approaches have also treated raindrops as a major degradation type within unified restoration networks, including rain, snow, haze, and raindrops, achieving strong general adverse-weather restoration performance as well as competitive raindrop removal results~\cite{li2020all,valanarasu2022transweather,qu2025weatherremover,sun2024restoring}. Diffusion-based methods have subsequently leveraged generative priors to restore images degraded by complex adverse-weather conditions, including raindrops~\cite{ozdenizci2023restoring,chen2024teaching}.

\subsection{Reflection Removal}
\label{sec:related-reflection-removal}

Early methods adopted CNNs to decompose the transmission and reflection layers~\cite{fan2017ceilnet,li2020ibcln}. Subsequent interaction-based methods modeled the transmission and reflection layers using separate feature streams~\cite{hu2021trash,hu2023dsrnet}. DSIT~\cite{hu2024single} further applied interactive attention to capture long-range relationships within and between the streams. Beyond these designs, RDNet~\cite{zhao2025reversible} employed a reversible encoder and transmission-rate-aware prompts to preserve information and adaptively separate reflection features. Diffusion-based methods later leveraged generative priors for complex reflection mixtures. L-DiffER~\cite{hong2024ldiffer} and DAI~\cite{hu2026dereflection} recovered transmission content using pretrained generative knowledge. Recently, FUMO~\cite{xu2026fumo} further incorporated reflection-intensity and high-frequency priors into a coarse-to-fine framework for spatially adaptive restoration.

\subsection{Unified Removal of Raindrops and Reflections (UR\textsuperscript{3})}
\label{sec:related-ur3}

Most previous restoration methods address raindrops and reflections separately. Simply cascading task-specific models does not fully consider how the two degradations interact. DiffUR\textsuperscript{3} first formulated the unified removal of raindrops and reflections (UR\textsuperscript{3}) task and introduced the real-world RDRF benchmark~\cite{liu2026diffur3}. It uses DRSformer to produce an initial restoration~\cite{chen2023drsformer}, followed by a Stable Diffusion-based stage conditioned on both the low-quality input and the initial result~\cite{rombach2022high}. Modulate\&Gate combines these conditions, and the Fidelity Encoder reduces VAE-induced structural distortions. However, when the underlying scene is largely hidden by raindrops or reflections, the diffusion prior may generate visually realistic content that does not match the actual scene. Our MAFNet-UR\textsuperscript{3} addresses this limitation through Stage III, which uses degradation masks to fuse complementary information from multiple observations and refine a scene-consistent restoration anchor.

\begin{figure}[t]
  \centering
  \resizebox{0.998\linewidth}{!}{%
  \includegraphics[width=\linewidth]{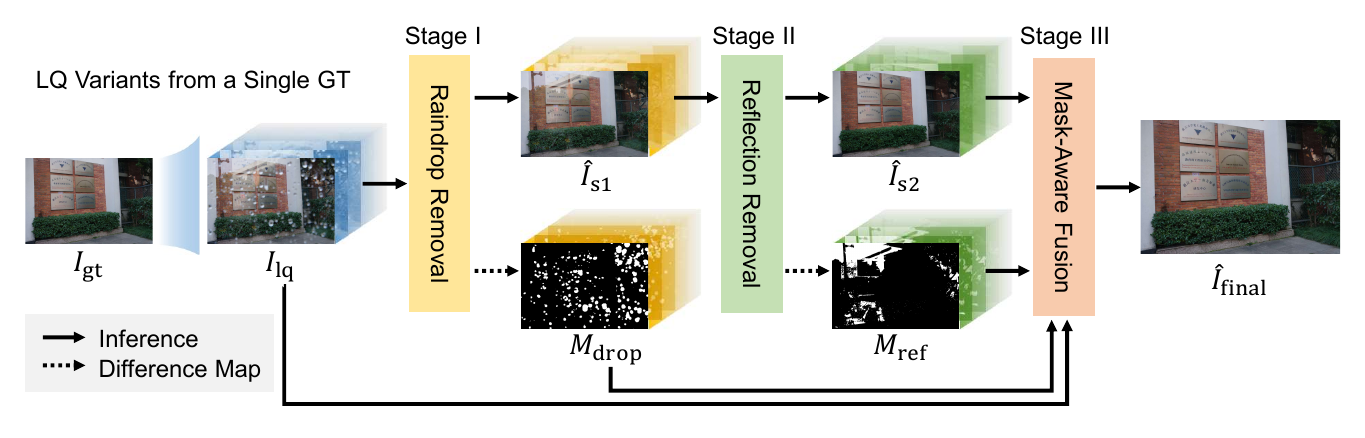}
  }
  \caption{Overview of the proposed MAFNet-UR\textsuperscript{3} framework with two observations ($N=2$). Stages I and II sequentially restore each observation, and Stage III performs mask-aware fusion across observations to produce $\hat{I}_{\mathrm{final}}$.}
  \label{fig:overall_pipeline}
\end{figure}

\section{Method}
\label{sec:method}

\subsection{Overall Three-Stage Framework}
\label{sec:overall-framework}

We propose a three-stage framework for unified raindrop and reflection removal. The input $I_{\mathrm{lq}}$ denotes the original low-quality image containing both raindrops and reflections. Stage I suppresses raindrops while retaining the reflection layer, producing $\hat{I}_{\mathrm{s1}}$. Stage II then performs prior-guided coarse-to-fine reflection removal on $\hat{I}_{\mathrm{s1}}$ to produce $\hat{I}_{\mathrm{s2}}$. Additionally, the raindrop mask $M_{\mathrm{drop}}$ and reflection mask $M_{\mathrm{ref}}$ are estimated by comparing the input image with the outputs of Stages I and II, respectively. Stage III then aggregates complementary predictions from multiple observations of the same scene to obtain the final fused output $\hat{I}_{\mathrm{final}}$. Figure~\ref{fig:overall_pipeline} illustrates the overall framework of the proposed method.

\subsection{Stage I: Reflection-Preserving Raindrop Removal}
\label{sec:stage-i}

Stage I only removes raindrops while preserving reflections for Stage II.
In particular, we fine-tune the pretrained WeatherRemover~\cite{qu2025weatherremover} using inputs that contain both raindrops and reflections. For each input $I_{\mathrm{lq}}$, the corresponding raindrop-free image $I_{\mathrm{s1}}$ that contains reflections is selected from the RDRF-training set as the supervision target.
By isolating raindrop removal in this manner, the supervision encourages Stage I to retain reflection content in $I_{\mathrm{s1}}$ for the subsequent reflection-removal stage. For training Stage I, we employ Charbonnier loss:
\begin{equation}
  \mathcal{L}_{\mathrm{stage1}}
  =\sqrt{\left\|I_{\mathrm{s1}}-\hat{I}_{\mathrm{s1}}\right\|_2^2+\epsilon^2}-\epsilon,
\end{equation}
where \(\hat{I}_{\mathrm{s1}}\) denotes the output of Stage I, \(\|\cdot\|_2\) denotes the L2 norm, and \(\epsilon\) is set to \(0.03\).

\subsection{Stage II: Prior-Guided Diffusion and Refinement}
\label{sec:stage-ii}

Stage II removes reflections from the Stage I output $\hat{I}_{\mathrm{s1}}$ while recovering transmission details. In particular, we adopt FUMO~\cite{xu2026fumo} to perform prior-guided coarse-to-fine restoration. Given $\hat{I}_{\mathrm{s1}}$, the diffusion module first produces an intermediate transmission result ${T}$, and the image-space refiner subsequently generates the Stage II output $\hat{I}_{\mathrm{s2}}$.

For prior guidance, we adopt two priors: a reflection-intensity prior, $P_{\mathrm{int}}$, and a high-frequency prior, $P_{\mathrm{hf}}$. In particular, the reflection-intensity prior $P_{\mathrm{int}}$ is generated from $\hat{I}_{\mathrm{s1}}$ using Qwen3-VL-8B-Instruct~\cite{bai2025qwen3} before training. This prior represents the severity of reflections and guides the model in determining the appropriate strength of reflection removal. We combine patch-level reflection strength estimates with global reflection localization and construct a dense prior map through localization-based enhancement, guided filtering, and normalization. We compute the high-frequency prior $P_{\mathrm{hf}}$ from $\hat{I}_{\mathrm{s1}}$ through multi-scale residual aggregation with channel-wise dilated smoothing at increasing scales. The resulting structure-sensitive prior guides the diffusion module and is also provided to the image-space refiner.

Concretely, we initialize our diffusion model with the pretrained weights of Stable Diffusion 2.1-base~\cite{rombach2022high} and incorporate a ControlNet-style branch~\cite{zhang2023adding} conditioned on $\hat{I}_{\mathrm{s1}}$. The two priors spatially modulate the ControlNet residuals as follows:
\begin{equation}
  g = 1 + \beta P_{\mathrm{int}} \odot P_{\mathrm{hf}},
\end{equation}
with $\beta = 0.25$, where \(P_{\mathrm{int}}\) and \(P_{\mathrm{hf}}\) highlight reflection-dominant regions and high-frequency details, respectively. We then use the gated conditioning for one-step diffusion restoration to generate ${T}$.

We train Stage II in two sequential phases. In the first phase, we optimize the coarse diffusion stage by updating only the ControlNet branch and the upsampling (decoder) blocks of the denoising U-Net, while keeping the VAE and the remaining U-Net parameters frozen.

Next, the NAFNet-based refiner~\cite{chen2022simple}, $R_\phi(\cdot)$, combines ${T}$, $P_{\mathrm{hf}}$, $P_{\mathrm{int}}$, and $\hat{I}_{\mathrm{s1}}$ to predict the refined clean output $\hat{I}_{\mathrm{s2}}$:
\begin{equation}
  \hat{I}_{\mathrm{s2}} = R_\phi(T, P_{\mathrm{hf}}, P_{\mathrm{int}}, \hat{I}_{\mathrm{s1}}).
\end{equation}
In the second phase, we freeze the entire trained diffusion model, including ControlNet and the U-Net, and optimize only the refiner using a weighted combination of a pixel loss, an LPIPS-based perceptual loss~\cite{zhang2018unreasonable}, and an edge-aware gradient loss~\cite{xu2026fumo}:
\begin{equation}
  \mathcal{L}_{\mathrm{refine}}=0.5\mathcal{L}_{1}+0.25\mathcal{L}_{\mathrm{LPIPS}}+0.25\mathcal{L}_{\mathrm{grad}},
\end{equation}
where $\mathcal{L}_{\mathrm{grad}}$ measures the discrepancy between the horizontal and vertical image gradients of the refined output and clean target using fixed Sobel filters, encouraging edge alignment and reducing local misalignment artifacts. For further details, please refer to the original FUMO paper~\cite{xu2026fumo}.

\subsection{Stage III: Mask-Aware Multi-Observation Fusion}
\label{sec:stage-iii}

\begin{figure}[t]
  \centering
  \includegraphics[width=\linewidth]{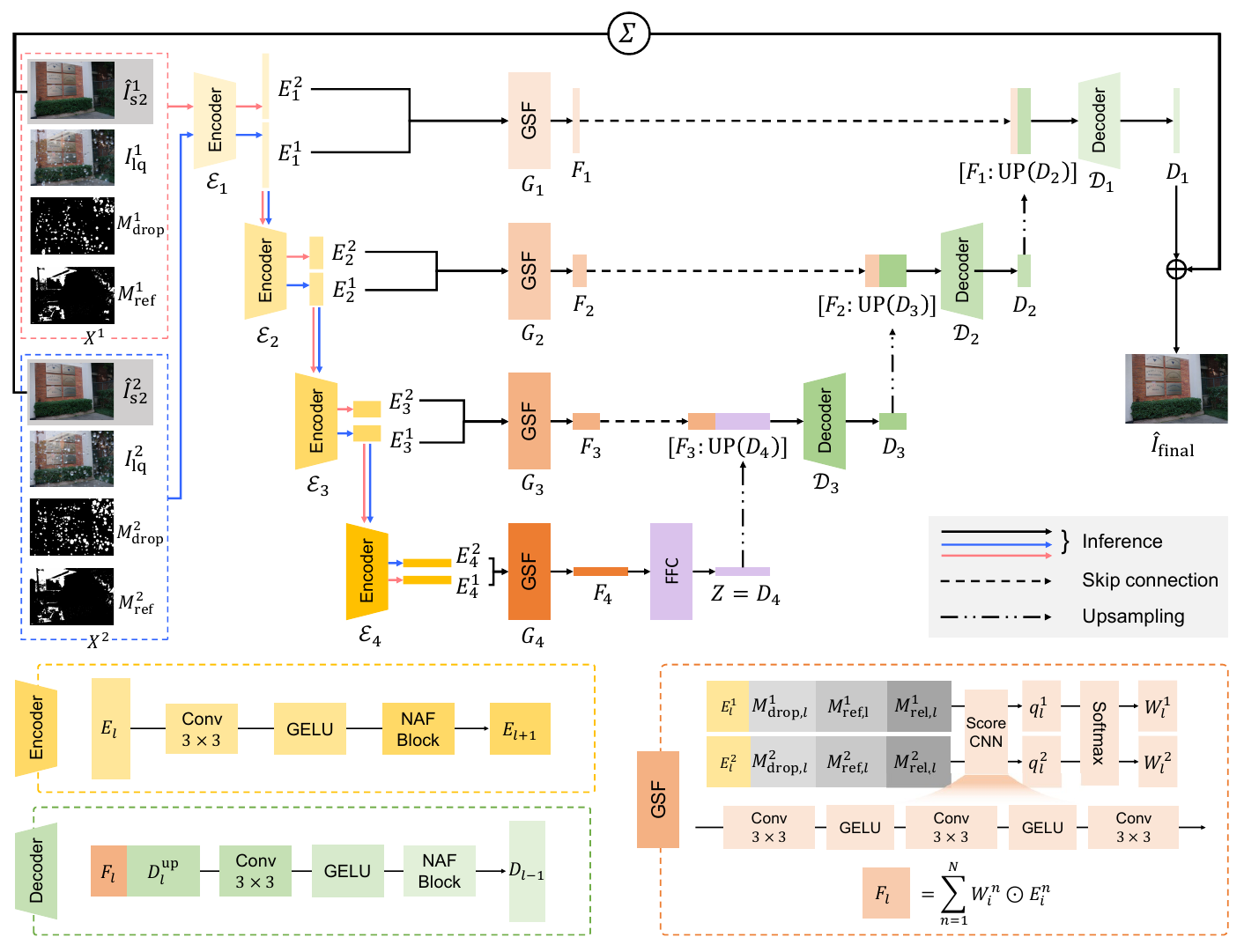}
  \caption{Stage III architecture for two observations ($N=2$). The main panel shows multi-scale encoding, GSF-based fusion across observations, and anchor-guided reconstruction. The bottom panels detail the encoder, decoder, and GSF modules.}
  \label{fig:stage3_architecture}
\end{figure}

Stage III performs mask-aware fusion across multiple observations of the same scene. Since raindrop and reflection artifacts vary across observations at each spatial location, we estimate spatial confidence and aggregate complementary features from relatively less-degraded regions in the Stage II outputs. Figure~\ref{fig:stage3_architecture} illustrates the overall architecture.

\paragraph{Mask Generation}
We first estimate two binary degradation masks for each observation, where $n$ denotes the observation index. Specifically, we compute the raindrop mask \(M_{\mathrm{drop}}^n\) by thresholding the luminance difference between the original low-quality image \(I_{\mathrm{lq}}^n\) and the Stage I output \(\hat{I}_{\mathrm{s1}}^n\). This mask retains regions whose luminance decreases after raindrop removal. Similarly, we obtain the reflection mask \(M_{\mathrm{ref}}^n\) by thresholding the luminance difference between \(\hat{I}_{\mathrm{s1}}^n\) and the Stage II output \(\hat{I}_{\mathrm{s2}}^n\), emphasizing regions whose luminance decreases after reflection removal. The masks are defined as
\begin{align}  
  M_{\mathrm{drop}}^n &= \mathbb{I}\left(L(I_{\mathrm{lq}}^n)-L(\hat{I}_{\mathrm{s1}}^n)>\tau_\mathrm{drop}\right),\\
  M_{\mathrm{ref}}^n &= \mathbb{I}\left(L(\hat{I}_{\mathrm{s1}}^n)-L(\hat{I}_{\mathrm{s2}}^n)>\tau_\mathrm{ref}\right),
\end{align}
where $L(\cdot)$ extracts the luminance (Y) channel in the YCbCr space, $\mathbb{I}(\cdot)$ denotes the indicator function, and $\tau_\mathrm{drop}$ and $\tau_\mathrm{ref}$ are the thresholds for each mask, respectively.

\paragraph{Encoder}
For each observation, we concatenate the Stage II output, original low-quality image, and two degradation masks along the channel dimension to generate an eight-channel input:
\begin{equation}
  X^n=[\hat{I}_{\mathrm{s2}}^n, I_{\mathrm{lq}}^n, M_{\mathrm{drop}}^n, M_{\mathrm{ref}}^n].
\end{equation}
A shared NAFBlock-based~\cite{chen2022simple} encoder takes each \(X^n\) and extracts features at four spatial scales. We denote the encoded feature of observation \(n\) at scale \(l\) as
\begin{equation}
  E_l^n=\mathcal{E}_l(X^n),
  \qquad l=1,2,3,4.
\end{equation}
Each successive scale is spatially downsampled by a factor of two. The resulting multi-scale features are then passed to the corresponding mask-aware fusion modules.

\paragraph{Gated Spatial Fusion}
For mask-aware fusion, we resize the raindrop and reflection masks to each feature scale with bilinear interpolation. This produces the soft mask cues \(M_{\mathrm{drop},l}^n\) and \(M_{\mathrm{ref},l}^n\). Then, their clipped sum and its complement are defined as the degradation cue and reliability cue, respectively:
\begin{align}
  M_{\mathrm{deg},l}^n&=\operatorname{clip}(M_{\mathrm{drop},l}^n+M_{\mathrm{ref},l}^n,0,1),\\
  M_{\mathrm{rel},l}^n&=1-M_{\mathrm{deg},l}^n.
\end{align}
At each scale, the Gated Spatial Fusion (GSF) module predicts spatial confidence from the encoded feature and mask cues. We normalize the confidence maps across the \(N\) observations at each spatial location and use the resulting weights to fuse their features:
\begin{align}
    q_l^n
    &=G_l\!\left(
      \left[
        E_l^n, M_{\mathrm{drop},l}^n, M_{\mathrm{ref},l}^n, M_{\mathrm{rel},l}^n
      \right]
    \right),\\
    W_l^n
    &=\frac{\exp(q_l^n)}{\sum_{j=1}^{N}\exp(q_l^j)},\\
    F_l
    &=\sum_{n=1}^{N} W_l^n \odot E_l^n,
\end{align}
where \(G_l(\cdot)\) denotes the GSF module at scale \(l\), \(q_l^n\) denotes the spatial confidence logit, and the softmax produces the normalized fusion weights \(W_l^n\), which satisfy \(\sum_{n=1}^{N}W_l^n=1\) at each spatial location.

\paragraph{Bottleneck and Decoder}
At the bottleneck, we refine the deepest fused feature \(F_4\) using residual Fast Fourier Convolution (FFC) blocks~\cite{chi2020fast}. These blocks combine local features with global frequency-domain filtering and apply a learnable channel-wise gate to each residual correction. We denote the resulting bottleneck feature as \(Z\). A NAFBlock-based decoder then reconstructs the output in a coarse-to-fine manner using the fused features as multi-scale skip connections:
\begin{align}
    D_4 &= Z,\\
    D_l
    &= \mathcal{D}_l\!\left(
      \left[F_l,\operatorname{UP}(D_{l+1})\right]
    \right),
\end{align}
where \(\mathcal{D}_l\) denotes the decoder block at scale \(l\), and \(\operatorname{UP}(\cdot)\) denotes bilinear upsampling. We use the finest-scale decoder feature \(D_1\) for residual reconstruction.

\paragraph{Reconstruction} Finally, Stage III performs anchor-based residual reconstruction instead of predicting the clean RGB image from scratch. We reuse the first-scale fusion weights to combine the Stage II outputs into a restoration anchor. From $D_1$, we predict the residual \(I_\mathrm{res}\) using a lightweight reconstruction head (Conv-GELU-Conv-tanh). We add this residual to the anchor as
\begin{align}
    \hat{I}_{\mathrm{anchor}}
    &= \sum_{n=1}^{N} W_1^n \odot \hat{I}_{\mathrm{s2}}^n,\\
    \hat{I}_{\mathrm{final}}
    &= \hat{I}_{\mathrm{anchor}} + s \, I_\mathrm{res},
\end{align}
where $s$ is the learnable residual scale.

For training Stage III, we use a weighted combination of pixel, structural, and LPIPS-based perceptual~\cite{zhang2018unreasonable} losses:
\begin{equation}
  \mathcal{L}_{\mathrm{stage3}}=\mathcal{L}_{1}+0.2\mathcal{L}_{\mathrm{SSIM}}+0.05\mathcal{L}_{\mathrm{LPIPS}}.
\end{equation}

\section{Experimental Results}
\label{sec:experimental-results}

\subsection{Implementation Details}
\label{sec:implementation-details}

\paragraph{Training and Evaluation Datasets}
We use the $1080 \times 720$ images provided in the training and testing splits of the RDRF benchmark~\cite{liu2026diffur3}. RDRF-training comprises 216 scenes with 9,003 image pairs, while RDRF-testing comprises 91 scenes with 277 image pairs. Each RDRF-testing scene contains one to five image pairs. All three stages are optimized on RDRF-training and evaluated on RDRF-testing.

\paragraph{Evaluation Metrics}

Following the evaluation protocol of DiffUR\textsuperscript{3}~\cite{liu2026diffur3}, we report six metrics: three full-reference metrics, PSNR, SSIM, and LPIPS~\cite{zhang2018unreasonable}, and three no-reference metrics, MUSIQ~\cite{ke2021musiq}, CLIP-IQA+~\cite{wang2023exploring}, and HyperIQA~\cite{su2020blindly}. Separately, we report the LoViF 2026 Challenge score, which combines LPIPS with PSNR and SSIM computed on the luminance channel $Y$:
\begin{equation}
  \mathrm{Score}=\mathrm{PSNR}(Y)+10\,\mathrm{SSIM}(Y)-5\,\mathrm{LPIPS}.
\end{equation}

\paragraph{Training Settings}
In Stage I, we optimize the raindrop removal model using Adam for 50k iterations with a constant learning rate of $5 \times 10^{-5}$. In Stage II, we first optimize the diffusion module using AdamW for 200k iterations with a learning rate of $2 \times 10^{-5}$. We then freeze the diffusion module and optimize the refinement module using AdamW for 25k iterations with a learning rate of $1 \times 10^{-4}$. In Stage III, we optimize MAFNet-UR\textsuperscript{3}-B and -L using AdamW with a learning rate of $1 \times 10^{-4}$. The B variant is trained for 100k iterations and contains approximately 38M Stage III parameters, whereas the L variant is trained for 400k iterations and contains approximately 96M. Both variants share the same Stage I and Stage II components. Unless otherwise specified, all experiments use two input observations ($N=2$). 

\paragraph{Observation Protocol}
We assume that multiple low-quality observations are available for each ground-truth image. During training, $N$ input observations are randomly sampled from the available observations of each scene. During inference, the observations are kept in their original order and partitioned into fixed, non-overlapping groups of $N$ observations. If the final group contains fewer than $N$ observations, the remaining input slots are zero-padded. 

Further implementation details are provided in the supplementary material.

\subsection{Quantitative Comparison}
\label{sec:quantitative-comparison}

Table~\ref{tab:quantitative-comparison} compares task-specific, cascaded, and unified methods on RDRF-testing. To further boost performance, we employ 8-way $D_4$ test-time augmentation (TTA) for our methods. For a fair comparison, the ranking is based on results without TTA. Note that methods with a ``+'' sign indicate the use of TTA.

\ExplSyntaxOn
\cs_new_protected:Npn \mafrank_store:nnn #1#2#3
  { \tl_gset:cn { g_mafrank_value_#1_#2_tl } {#3} }

\cs_new_protected:Npn \mafrank_add:nnn #1#2#3
  {
    \seq_if_exist:cF { g_mafrank_values_#1_#2_seq }
      { \seq_new:c { g_mafrank_values_#1_#2_seq } }
    \seq_gput_right:cn { g_mafrank_values_#1_#2_seq } {#3}
  }

\cs_new_protected:Npn \mafrank_declare_one:nnnnnnnn #1#2#3#4#5#6#7#8
  {
    \mafrank_store:nnn { table1_#1 } { psnr  } {#2}
    \mafrank_store:nnn { table1_#1 } { ssim  } {#3}
    \mafrank_store:nnn { table1_#1 } { lpips } {#4}
    \mafrank_store:nnn { table1_#1 } { musiq } {#5}
    \mafrank_store:nnn { table1_#1 } { clip  } {#6}
    \mafrank_store:nnn { table1_#1 } { hyper } {#7}
    \mafrank_store:nnn { table1_#1 } { score } {#8}
    \mafrank_add:nnn { table1 } { psnr  } {#2}
    \mafrank_add:nnn { table1 } { ssim  } {#3}
    \mafrank_add:nnn { table1 } { lpips } {#4}
    \mafrank_add:nnn { table1 } { musiq } {#5}
    \mafrank_add:nnn { table1 } { clip  } {#6}
    \mafrank_add:nnn { table1 } { hyper } {#7}
    \mafrank_add:nnn { table1 } { score } {#8}
  }

\cs_new_protected:Npn \mafrank_declare_one_unranked:nnnnnnnn #1#2#3#4#5#6#7#8
  {
    \mafrank_store:nnn { table1_#1 } { psnr  } {#2}
    \mafrank_store:nnn { table1_#1 } { ssim  } {#3}
    \mafrank_store:nnn { table1_#1 } { lpips } {#4}
    \mafrank_store:nnn { table1_#1 } { musiq } {#5}
    \mafrank_store:nnn { table1_#1 } { clip  } {#6}
    \mafrank_store:nnn { table1_#1 } { hyper } {#7}
    \mafrank_store:nnn { table1_#1 } { score } {#8}
  }

\cs_new_protected:Npn \mafrank_declare_two:nnnnnn #1#2#3#4#5#6
  {
    \mafrank_store:nnn { t2_#1_#2 } { psnr  } {#3}
    \mafrank_store:nnn { t2_#1_#2 } { ssim  } {#4}
    \mafrank_store:nnn { t2_#1_#2 } { lpips } {#5}
    \mafrank_store:nnn { t2_#1_#2 } { score } {#6}
    \mafrank_add:nnn { t2_#1 } { psnr  } {#3}
    \mafrank_add:nnn { t2_#1 } { ssim  } {#4}
    \mafrank_add:nnn { t2_#1 } { lpips } {#5}
    \mafrank_add:nnn { t2_#1 } { score } {#6}
  }

\cs_new_protected:Npn \mafrank_paint:nn #1#2
  {
    \int_case:nnF {#1}
      {
        {1}{\group_begin:\setlength{\fboxsep}{1pt}\colorbox[HTML]{F4CCCC}{\strut #2}\group_end:}
        {2}{\group_begin:\setlength{\fboxsep}{1pt}\colorbox[HTML]{FCE5CD}{\strut #2}\group_end:}
        {3}{\group_begin:\setlength{\fboxsep}{1pt}\colorbox[HTML]{FFF2CC}{\strut #2}\group_end:}
      }
      {#2}
  }

\cs_new_protected:Npn \mafrank_cell:nnnn #1#2#3#4
  {
    \tl_set:Nx \l_tmpa_tl { \tl_use:c { g_mafrank_value_#1_#2_#3_tl } }
    \int_zero:N \l_tmpa_int
    \seq_map_inline:cn { g_mafrank_values_#1_#3_seq }
      {
        \str_if_eq:nnTF {#4} { max }
          { \fp_compare:nT { ##1 > \l_tmpa_tl } { \int_incr:N \l_tmpa_int } }
          { \fp_compare:nT { ##1 < \l_tmpa_tl } { \int_incr:N \l_tmpa_int } }
      }
    \mafrank_paint:nn { \int_eval:n { \l_tmpa_int + 1 } } { \l_tmpa_tl }
  }

\NewDocumentCommand{\MAFRankOneData}{mmmmmmmm}
  { \mafrank_declare_one:nnnnnnnn {#1}{#2}{#3}{#4}{#5}{#6}{#7}{#8} }
\NewDocumentCommand{\MAFRankOneUnrankedData}{mmmmmmmm}
  { \mafrank_declare_one_unranked:nnnnnnnn {#1}{#2}{#3}{#4}{#5}{#6}{#7}{#8} }
\NewDocumentCommand{\MAFRankTwoData}{mmmmmm}
  { \mafrank_declare_two:nnnnnn {#1}{#2}{#3}{#4}{#5}{#6} }
\NewDocumentCommand{\MAFRankCell}{mmmm}
  { \mafrank_cell:nnnn {#1}{#2}{#3}{#4} }
\NewDocumentCommand{\MAFRankPlain}{mmm}
  { \tl_use:c { g_mafrank_value_#1_#2_#3_tl } }
\ExplSyntaxOff

\MAFRankOneData{agan}       {25.28}{0.9149}{0.1168}{72.92}{0.6407}{0.6398}{33.8500}
\MAFRankOneData{histo}      {25.74}{0.9217}{0.1214}{72.35}{0.5956}{0.6465}{34.3494}
\MAFRankOneData{rdnet}      {26.76}{0.9166}{0.1275}{70.69}{0.5976}{0.6259}{35.2838}
\MAFRankOneData{dsit}       {26.47}{0.9131}{0.1334}{71.29}{0.6047}{0.6248}{34.9383}
\MAFRankOneData{dai}        {27.48}{0.9251}{0.1058}{73.25}{0.6335}{0.6518}{36.2021}
\MAFRankOneData{agan-dsit}  {26.65}{0.9228}{0.1021}{73.58}{0.6782}{0.6526}{35.3680}
\MAFRankOneData{dsit-agan}  {26.56}{0.9184}{0.1061}{73.53}{0.6633}{0.6463}{35.2133}
\MAFRankOneData{agan-rdnet} {26.44}{0.8979}{0.1297}{71.90}{0.6613}{0.5759}{34.7691}
\MAFRankOneData{rdnet-agan} {26.63}{0.9205}{0.1024}{73.25}{0.6583}{0.6489}{35.3188}
\MAFRankOneData{histo-dai}  {27.88}{0.9350}{0.0924}{74.10}{0.6417}{0.6745}{36.7650}
\MAFRankOneData{dai-histo}  {27.65}{0.9296}{0.0943}{73.79}{0.6441}{0.6628}{36.4786}
\MAFRankOneData{diffur}     {29.41}{0.9372}{0.0813}{74.72}{0.6705}{0.7046}{38.3755}
\MAFRankOneData{stage1}     {26.21}{0.9266}{0.1062}{72.58}{0.6289}{0.6566}{34.9490}
\MAFRankOneData{stage12}    {30.09}{0.9369}{0.0771}{72.45}{0.6463}{0.6548}{39.0752}
\MAFRankOneData{maf-b}      {31.34}{0.9558}{0.0483}{73.51}{0.6771}{0.6844}{40.6543}
\MAFRankOneData{maf-l}      {31.74}{0.9608}{0.0415}{74.01}{0.6796}{0.7023}{41.1389}
\MAFRankOneUnrankedData{maf-b-tta}{31.60}{0.9577}{0.0473}{73.59}{0.6754}{0.6857}{40.9413}
\MAFRankOneUnrankedData{maf-l-tta}{31.95}{0.9621}{0.0413}{74.06}{0.6780}{0.7021}{41.3694}

\begin{table}[t]
  \centering 
  \caption{Quantitative comparison on RDRF-testing. The first-, second-, and third-best results for each metric are highlighted in \highlightr{red}, \highlighto{orange}, and \highlighty{yellow}, respectively.}
  \label{tab:quantitative-comparison}
  \scriptsize
  \setlength{\tabcolsep}{1.5pt}
  \resizebox{\linewidth}{!}{%
  \begin{tabular}{llccccccc}
    \toprule
    & & \multicolumn{3}{c}{Full-reference} & \multicolumn{3}{c}{No-reference} & \\
    \cmidrule(lr){3-5} \cmidrule(lr){6-8}
    Task & Method & PSNR$\uparrow$ & SSIM$\uparrow$ & LPIPS$\downarrow$ & MUSIQ$\uparrow$ & CLIP-IQA+$\uparrow$ & HyperIQA$\uparrow$ & Score$\uparrow$ \\
    \midrule
    \multirow{2}{*}{\shortstack[l]{Raindrop\\removal}}
      & AGAN~\cite{qian2018attentive}       & \MAFRankCell{table1}{agan}{psnr}{max} & \MAFRankCell{table1}{agan}{ssim}{max} & \MAFRankCell{table1}{agan}{lpips}{min} & \MAFRankCell{table1}{agan}{musiq}{max} & \MAFRankCell{table1}{agan}{clip}{max} & \MAFRankCell{table1}{agan}{hyper}{max} & \MAFRankCell{table1}{agan}{score}{max} \\
      & Histoformer~\cite{sun2024restoring} & \MAFRankCell{table1}{histo}{psnr}{max} & \MAFRankCell{table1}{histo}{ssim}{max} & \MAFRankCell{table1}{histo}{lpips}{min} & \MAFRankCell{table1}{histo}{musiq}{max} & \MAFRankCell{table1}{histo}{clip}{max} & \MAFRankCell{table1}{histo}{hyper}{max} & \MAFRankCell{table1}{histo}{score}{max} \\
    \midrule
    \multirow{3}{*}{\shortstack[l]{Reflection\\removal}}
      & RDNet~\cite{zhao2025reversible} & \MAFRankCell{table1}{rdnet}{psnr}{max} & \MAFRankCell{table1}{rdnet}{ssim}{max} & \MAFRankCell{table1}{rdnet}{lpips}{min} & \MAFRankCell{table1}{rdnet}{musiq}{max} & \MAFRankCell{table1}{rdnet}{clip}{max} & \MAFRankCell{table1}{rdnet}{hyper}{max} & \MAFRankCell{table1}{rdnet}{score}{max} \\
      & DSIT~\cite{hu2024single}          & \MAFRankCell{table1}{dsit}{psnr}{max} & \MAFRankCell{table1}{dsit}{ssim}{max} & \MAFRankCell{table1}{dsit}{lpips}{min} & \MAFRankCell{table1}{dsit}{musiq}{max} & \MAFRankCell{table1}{dsit}{clip}{max} & \MAFRankCell{table1}{dsit}{hyper}{max} & \MAFRankCell{table1}{dsit}{score}{max} \\
      & DAI~\cite{hu2026dereflection}            & \MAFRankCell{table1}{dai}{psnr}{max} & \MAFRankCell{table1}{dai}{ssim}{max} & \MAFRankCell{table1}{dai}{lpips}{min} & \MAFRankCell{table1}{dai}{musiq}{max} & \MAFRankCell{table1}{dai}{clip}{max} & \MAFRankCell{table1}{dai}{hyper}{max} & \MAFRankCell{table1}{dai}{score}{max} \\
    \midrule
    \multirow{6}{*}{Cascade}
      & AGAN $\rightarrow$ DSIT       & \MAFRankCell{table1}{agan-dsit}{psnr}{max} & \MAFRankCell{table1}{agan-dsit}{ssim}{max} & \MAFRankCell{table1}{agan-dsit}{lpips}{min} & \MAFRankCell{table1}{agan-dsit}{musiq}{max} & \MAFRankCell{table1}{agan-dsit}{clip}{max} & \MAFRankCell{table1}{agan-dsit}{hyper}{max} & \MAFRankCell{table1}{agan-dsit}{score}{max} \\
      & DSIT $\rightarrow$ AGAN       & \MAFRankCell{table1}{dsit-agan}{psnr}{max} & \MAFRankCell{table1}{dsit-agan}{ssim}{max} & \MAFRankCell{table1}{dsit-agan}{lpips}{min} & \MAFRankCell{table1}{dsit-agan}{musiq}{max} & \MAFRankCell{table1}{dsit-agan}{clip}{max} & \MAFRankCell{table1}{dsit-agan}{hyper}{max} & \MAFRankCell{table1}{dsit-agan}{score}{max} \\
      & AGAN $\rightarrow$ RDNet      & \MAFRankCell{table1}{agan-rdnet}{psnr}{max} & \MAFRankCell{table1}{agan-rdnet}{ssim}{max} & \MAFRankCell{table1}{agan-rdnet}{lpips}{min} & \MAFRankCell{table1}{agan-rdnet}{musiq}{max} & \MAFRankCell{table1}{agan-rdnet}{clip}{max} & \MAFRankCell{table1}{agan-rdnet}{hyper}{max} & \MAFRankCell{table1}{agan-rdnet}{score}{max} \\
      & RDNet $\rightarrow$ AGAN      & \MAFRankCell{table1}{rdnet-agan}{psnr}{max} & \MAFRankCell{table1}{rdnet-agan}{ssim}{max} & \MAFRankCell{table1}{rdnet-agan}{lpips}{min} & \MAFRankCell{table1}{rdnet-agan}{musiq}{max} & \MAFRankCell{table1}{rdnet-agan}{clip}{max} & \MAFRankCell{table1}{rdnet-agan}{hyper}{max} & \MAFRankCell{table1}{rdnet-agan}{score}{max} \\
      & Histoformer $\rightarrow$ DAI & \MAFRankCell{table1}{histo-dai}{psnr}{max} & \MAFRankCell{table1}{histo-dai}{ssim}{max} & \MAFRankCell{table1}{histo-dai}{lpips}{min} & \MAFRankCell{table1}{histo-dai}{musiq}{max} & \MAFRankCell{table1}{histo-dai}{clip}{max} & \MAFRankCell{table1}{histo-dai}{hyper}{max} & \MAFRankCell{table1}{histo-dai}{score}{max} \\
      & DAI $\rightarrow$ Histoformer & \MAFRankCell{table1}{dai-histo}{psnr}{max} & \MAFRankCell{table1}{dai-histo}{ssim}{max} & \MAFRankCell{table1}{dai-histo}{lpips}{min} & \MAFRankCell{table1}{dai-histo}{musiq}{max} & \MAFRankCell{table1}{dai-histo}{clip}{max} & \MAFRankCell{table1}{dai-histo}{hyper}{max} & \MAFRankCell{table1}{dai-histo}{score}{max} \\
    \midrule
    UR\textsuperscript{3}
      & DiffUR\textsuperscript{3}\textsuperscript{$\dagger$}~\cite{liu2026diffur3} & \MAFRankCell{table1}{diffur}{psnr}{max} & \MAFRankCell{table1}{diffur}{ssim}{max} & \MAFRankCell{table1}{diffur}{lpips}{min} & \MAFRankCell{table1}{diffur}{musiq}{max} & \MAFRankCell{table1}{diffur}{clip}{max} & \MAFRankCell{table1}{diffur}{hyper}{max} & \MAFRankCell{table1}{diffur}{score}{max} \\
    \midrule
    \multirow{6}{*}{Ours}
      & Stage I                        & \MAFRankCell{table1}{stage1}{psnr}{max} & \MAFRankCell{table1}{stage1}{ssim}{max} & \MAFRankCell{table1}{stage1}{lpips}{min} & \MAFRankCell{table1}{stage1}{musiq}{max} & \MAFRankCell{table1}{stage1}{clip}{max} & \MAFRankCell{table1}{stage1}{hyper}{max} & \MAFRankCell{table1}{stage1}{score}{max} \\
      & Stage I+II                     & \MAFRankCell{table1}{stage12}{psnr}{max} & \MAFRankCell{table1}{stage12}{ssim}{max} & \MAFRankCell{table1}{stage12}{lpips}{min} & \MAFRankCell{table1}{stage12}{musiq}{max} & \MAFRankCell{table1}{stage12}{clip}{max} & \MAFRankCell{table1}{stage12}{hyper}{max} & \MAFRankCell{table1}{stage12}{score}{max} \\
      & MAFNet-UR\textsuperscript{3}-B & \MAFRankCell{table1}{maf-b}{psnr}{max} & \MAFRankCell{table1}{maf-b}{ssim}{max} & \MAFRankCell{table1}{maf-b}{lpips}{min} & \MAFRankCell{table1}{maf-b}{musiq}{max} & \MAFRankCell{table1}{maf-b}{clip}{max} & \MAFRankCell{table1}{maf-b}{hyper}{max} & \MAFRankCell{table1}{maf-b}{score}{max} \\
      & MAFNet-UR\textsuperscript{3}-L & \MAFRankCell{table1}{maf-l}{psnr}{max} & \MAFRankCell{table1}{maf-l}{ssim}{max} & \MAFRankCell{table1}{maf-l}{lpips}{min} & \MAFRankCell{table1}{maf-l}{musiq}{max} & \MAFRankCell{table1}{maf-l}{clip}{max} & \MAFRankCell{table1}{maf-l}{hyper}{max} & \MAFRankCell{table1}{maf-l}{score}{max} \\
    \cmidrule(l){2-9}
      & MAFNet-UR\textsuperscript{3}-B{+} & \MAFRankPlain{table1}{maf-b-tta}{psnr} & \MAFRankPlain{table1}{maf-b-tta}{ssim} & \MAFRankPlain{table1}{maf-b-tta}{lpips} & \MAFRankPlain{table1}{maf-b-tta}{musiq} & \MAFRankPlain{table1}{maf-b-tta}{clip} & \MAFRankPlain{table1}{maf-b-tta}{hyper} & \MAFRankPlain{table1}{maf-b-tta}{score} \\
      & MAFNet-UR\textsuperscript{3}-L{+} & \MAFRankPlain{table1}{maf-l-tta}{psnr} & \MAFRankPlain{table1}{maf-l-tta}{ssim} & \MAFRankPlain{table1}{maf-l-tta}{lpips} & \MAFRankPlain{table1}{maf-l-tta}{musiq} & \MAFRankPlain{table1}{maf-l-tta}{clip} & \MAFRankPlain{table1}{maf-l-tta}{hyper} & \MAFRankPlain{table1}{maf-l-tta}{score} \\
    \bottomrule
  \end{tabular}
  }
  \vspace{2pt}

  \parbox{\linewidth}{\scriptsize \textsuperscript{$\dagger$} Results are taken from the original paper because the official implementation and pretrained weights are not publicly available.}
\end{table}

\paragraph{Comparison with Existing Methods}
Models designed solely for raindrop or reflection removal show limited performance when both degradations are present. Cascading two task-specific models can improve upon their individual use, confirming the potential benefit of addressing both degradations sequentially, but a considerable performance gap remains. MAFNet-UR\textsuperscript{3}-L outperforms the individual models and their cascades on all evaluation metrics except MUSIQ, where it remains competitive. These results demonstrate the benefit of handling both degradations within a coordinated framework.

We also compare our method with the recent DiffUR\textsuperscript{3}, the state-of-the-art method for the UR\textsuperscript{3} task.
MAFNet-UR\textsuperscript{3}-L achieves superior results across all full-reference metrics and the challenge score. For the no-reference metrics, it performs best on CLIP-IQA+ and remains comparable on MUSIQ and HyperIQA.

\paragraph{Effect of Multi-Observation Restoration}
The effect of Stage III is also evident. Relative to the Stage I+II output, MAFNet-UR\textsuperscript{3}-L improves PSNR by 1.65 dB and SSIM by 0.0239 while reducing LPIPS by 0.0356. These gains indicate that mask-aware multi-observation fusion helps correct artifacts remaining after the preceding stages and improves reconstruction accuracy.

\paragraph{Effect of Test-Time Augmentation}
TTA consistently improves the full-reference metrics for both model sizes, increasing PSNR and SSIM while reducing LPIPS. Its effects on the no-reference metrics are mixed, suggesting that geometric TTA ensembling primarily benefits full-reference reconstruction accuracy. On RDRF-testing, MAFNet-UR\textsuperscript{3}-L{+} achieves state-of-the-art performance, with a PSNR of 31.95 dB, an SSIM of 0.9621, and an LPIPS of 0.0413. We submitted this configuration to the LoViF 2026 UR\textsuperscript{3} Challenge, where it achieved a score of 41.1416 and ranked \textbf{third} on the challenge test set.

\subsection{Qualitative Comparison}
\label{sec:qualitative-comparison}

\paragraph{Comparison with Existing Methods}

\begin{figure}[!t]
  \centering
  \resizebox{0.96\linewidth}{!}{%
    \includegraphics[width=\linewidth]{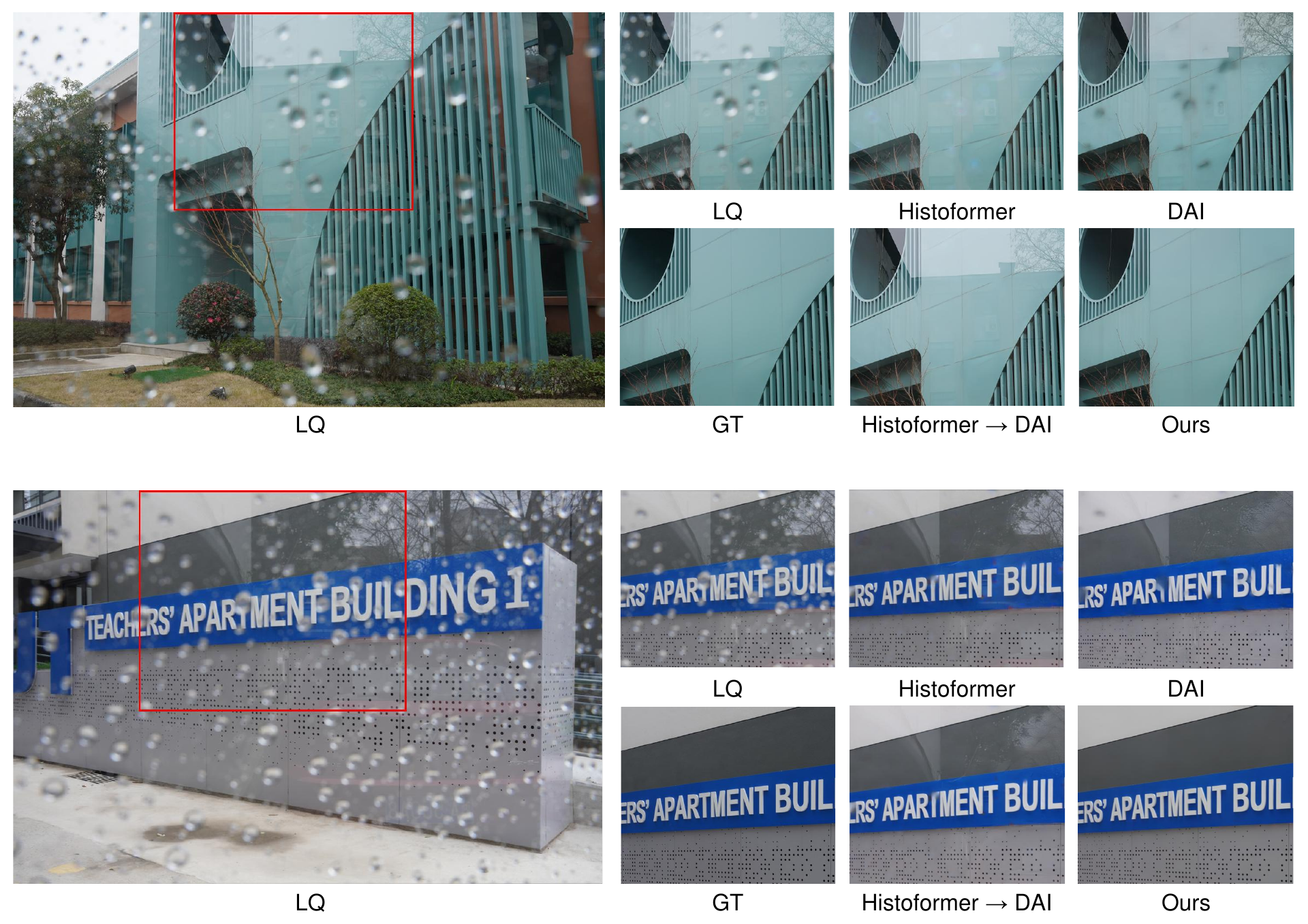}
  }
  \vspace{-5pt}
  \caption{Qualitative comparisons on RDRF-testing.}
  \vspace{-5pt}
  \label{fig:qualitative-comparison}
\end{figure}

Figure~\ref{fig:qualitative-comparison} compares task-specific models, their cascade, and our results on cropped regions.
As shown, Histoformer~\cite{sun2024restoring} effectively suppresses raindrops, but substantial reflection content remains, whereas DAI~\cite{hu2026dereflection} reduces reflections while leaving raindrop-corrupted regions insufficiently restored. The cascaded Histoformer $\rightarrow$ DAI alleviates both degradations to some extent, but residual raindrops and reflections remain in heavily corrupted regions. The cascade also introduces local blur, oversmoothing, and texture loss, indicating that simply connecting independently trained task-specific models fails to fully recover information lost or distorted in the preceding stage.

In contrast, our method successfully removes both degradations and better preserves scene boundaries, fine textures, and local structures. By leveraging degradation cues and encoded image features, the mask-aware fusion module integrates complementary information across observations. This process suppresses residual degradations and artifacts introduced by the preceding stages, yielding cleaner and more scene-consistent restorations.

\begin{figure}[!t]
  \centering
  \includegraphics[width=\linewidth]{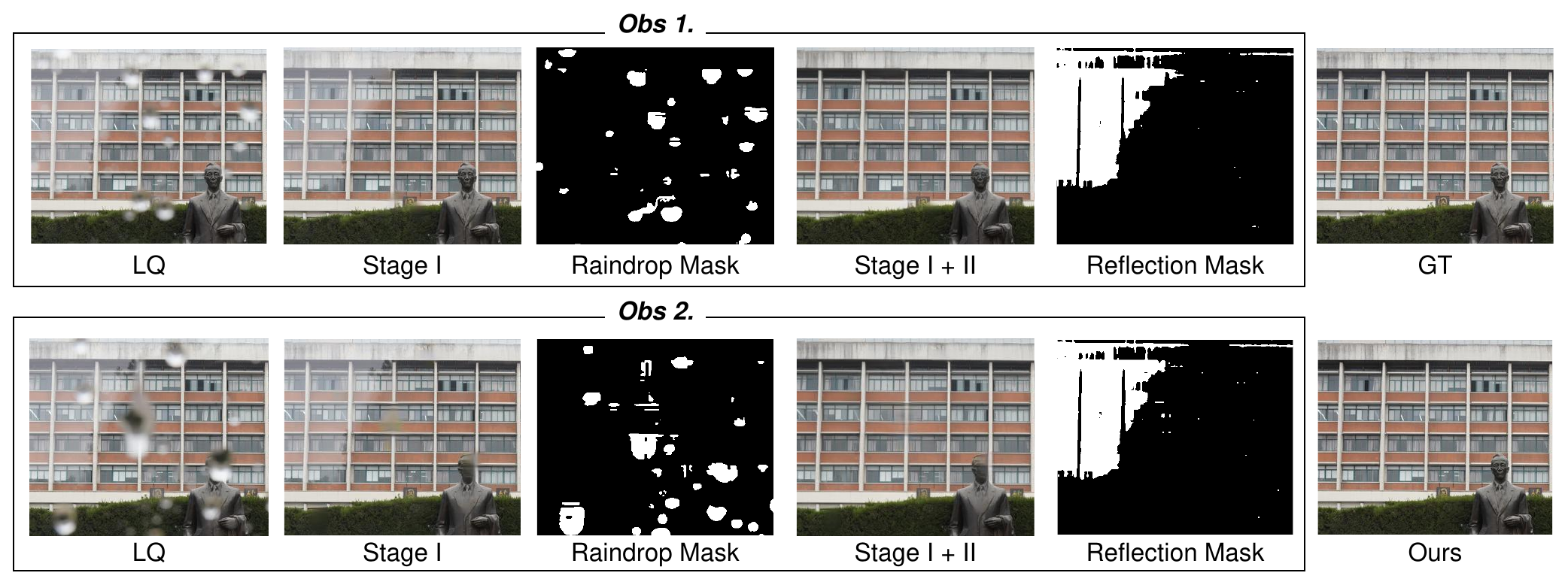}
  \caption{Stage-wise multi-observation restoration ($N=2$). The boxed regions show the restoration results for Observations 1 and 2. The ground truth (top) and final output (bottom) are displayed separately in the rightmost column.}
  \label{fig:multi-observation-restoration}
\end{figure}

\paragraph{Stage-Wise Comparisons of Multi-Observation Restoration}
Figure~\ref{fig:multi-observation-restoration} presents the complete restoration results for two observations ($N=2$), showing their low-quality inputs and corresponding Stage I and Stage II outputs, followed by the final mask-aware fusion result from Stage III.

Stage I removes most raindrops while largely preserving the reflection layer, and the subsequent Stage II reduces the remaining reflection content. However, in severely occluded regions with limited visual evidence, the diffusion prior in Stage II may generate visually plausible but scene-inconsistent content. Moreover, because the locations and severity of the degradations vary across observations, the Stage I+II results exhibit different local errors.

The complete three-stage framework produces a cleaner and more consistent result by exploiting these complementary predictions. Stage III uses the degradation cues and encoded image features to determine which observation is more reliable at each location. It then fuses the corresponding features, reducing artifacts from the preceding stages and preserving scene structure.

\section{Analysis and Discussion}
\label{sec:analysis-discussion}

\subsection{Ablation Study}
\label{sec:ablation-study}

Table~\ref{tab:stage3-ablation} summarizes the ablation results for the Stage III inputs, GSF module, degradation mask representation, and observation count. Unless otherwise specified, experiments use MAFNet-UR\textsuperscript{3}-B with \(N=2\) (ours). Additional experimental settings are provided in the supplementary material.

\begin{table}[t]
  \centering
  \caption{Ablation study of Stage III. The best result within each group is highlighted in \textbf{bold}.}
  \label{tab:stage3-ablation}
  \footnotesize
  \setlength{\tabcolsep}{1.5pt}
  \begin{tabular}{llcccc}
    \toprule
    Experiment & Configuration & PSNR$\uparrow$ & SSIM$\uparrow$ & LPIPS$\downarrow$ & Score$\uparrow$ \\
    \midrule
    \multirow{3}{*}{\shortstack[l]{Input\\composition}}
      & w/o $I_{\mathrm{lq}}$                     & 30.79 & 0.9498 & 0.0536 & 40.0200 \\
      & w/o degradation masks                     & 31.32 & 0.9557 & 0.0487 & 40.6305 \\
      & w/ $I_{\mathrm{lq}}$ and masks (ours) & \textbf{31.34} & \textbf{0.9558} & \textbf{0.0483} & \textbf{40.6543} \\
    \midrule
    \multirow{2}{*}{\shortstack[l]{GSF\\module}}
      & Handcrafted mask weighting                 & 31.27 & 0.9553 & 0.0497 & 40.5733 \\
      & GSF module (ours)                      & \textbf{31.34} & \textbf{0.9558} & \textbf{0.0483} & \textbf{40.6543} \\
    \midrule
    \multirow{4}{*}{\shortstack[l]{Mask\\representation}}
      & Luminance absolute difference             & 31.33 & 0.9557 & 0.0487 & 40.6440 \\
      & Otsu-thresholded luminance                & 31.33 & \textbf{0.9559} & \textbf{0.0483} & 40.6489 \\
      & RGB absolute difference                   & 31.33 & \textbf{0.9559} & 0.0484 & 40.6513 \\
      & Positive luminance difference (ours)  & \textbf{31.34} & 0.9558 & \textbf{0.0483} & \textbf{40.6543} \\
    \midrule
    \multirow{5}{*}{\shortstack[l]{Number of\\observations}}
      & $N=1$                                      & 30.50 & 0.9466 & 0.0573 & 39.6774 \\
      & $N=2$ (ours)                           & 31.34 & 0.9558 & 0.0483 & 40.6543 \\
      & $N=3$                                      & 31.89 & 0.9612 & 0.0431 & 41.2813 \\
      & $N=4$                                      & 31.98 & 0.9618 & 0.0429 & 41.3885 \\
      & $N=5$                                      & \textbf{32.14} & \textbf{0.9637} & \textbf{0.0410} & \textbf{41.5755} \\
    \bottomrule
  \end{tabular}
\end{table}

\paragraph{Input Compositions}
Adding the original low-quality image $I_{\mathrm{lq}}$ improves PSNR by 0.55 dB and SSIM by 0.0060, while reducing LPIPS by 0.0053. Introducing the degradation masks provides further improvements. These results show that $I_{\mathrm{lq}}$ provides useful scene and degradation cues beyond the Stage II output and that explicit spatial degradation cues further benefit multi-observation fusion.

\paragraph{GSF Module}
To demonstrate the effectiveness of the GSF module, we compare it with a straightforward fusion method that directly applies a weighted sum with the combined degradation masks. As shown in \tablename~\ref{tab:stage3-ablation}, using the GSF module instead of a handcrafted weighting scheme improves all evaluation metrics. These results indicate that mask information alone is insufficient for estimating fusion confidence and that encoded visual features provide complementary cues.

\paragraph{Mask Representations}
Across the four representations, the differences are negligible, with PSNR, SSIM, and LPIPS varying by at most 0.01 dB, 0.0002, and 0.0004, respectively. Together with the input-composition results, these small differences show that using degradation masks is more effective than omitting them, while performance is relatively insensitive to the specific mask representation. We adopt the fixed-threshold positive luminance difference because it achieves the highest PSNR and the joint-lowest LPIPS while remaining the simplest representation. Detailed mask definitions are provided in the supplementary material.

\paragraph{Number of Observations}
All configurations with $N \in \{1, 2, 3, 4, 5\}$ are evaluated on the same full RDRF-testing set, without excluding scenes based on the number of available observations. Performance improves monotonically across all metrics as $N$ increases from 1 to 5. Over this range, PSNR and SSIM improve by 1.64 dB and 0.0171, respectively, while LPIPS decreases by 0.0163. The largest gain occurs when increasing from $N=1$ to $N=2$, while additional observations provide further improvements, confirming that the framework can exploit a variable number of complementary observations.

\subsection{Discussion}
\label{sec:discussion}

\begin{figure}[t]
  \centering
  \resizebox{0.92\linewidth}{!}{%
  \includegraphics[width=\linewidth]{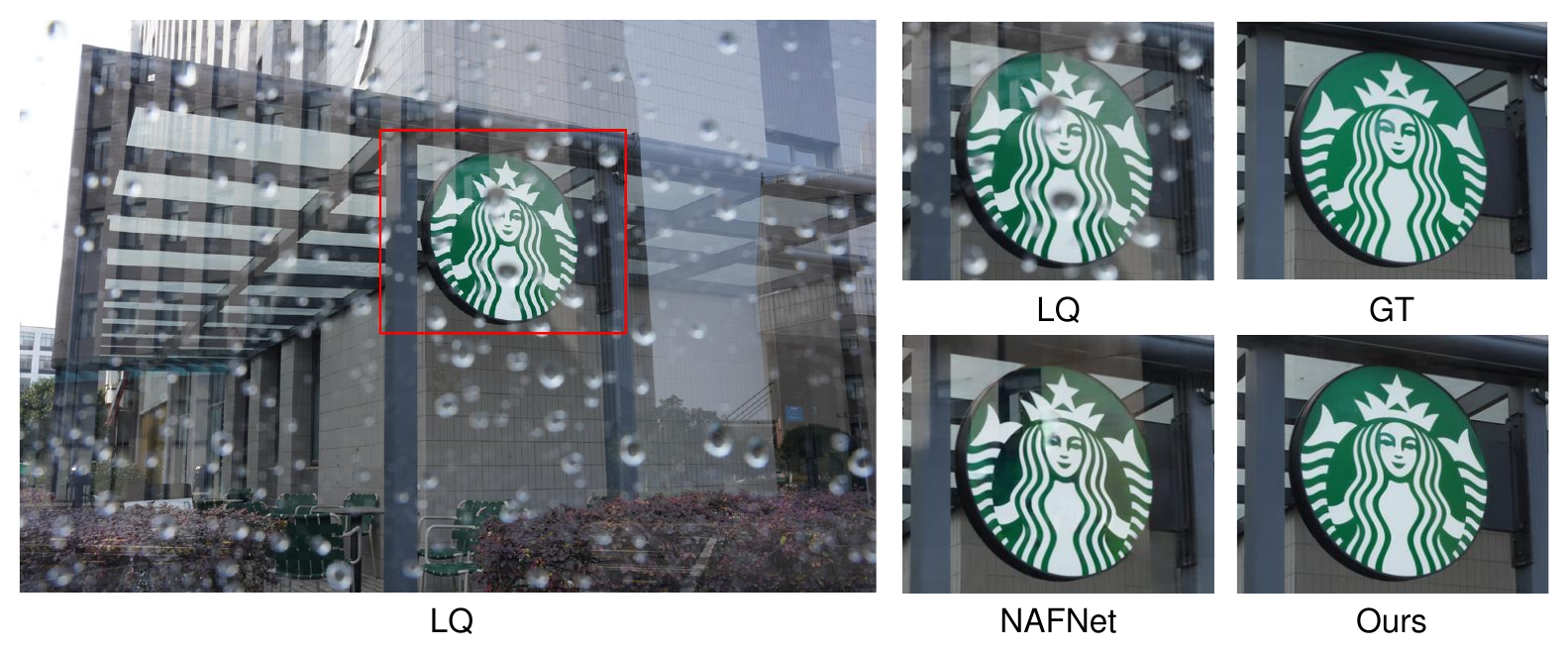}
  }
  \caption{Single-stage versus stage-wise restoration.}

  \label{fig:discussion-stagewise}
\end{figure}

\paragraph{Comparison of Stage-wise and Unified Restoration}
We compare the proposed stage-wise framework against a single-stage method. In particular, we employ NAFNet~\cite{chen2022simple} as an end-to-end unified comparison, configured with 64 channels and trained to directly predict the clean target from the low-quality input. Figure~\ref{fig:discussion-stagewise} presents the corresponding qualitative comparisons.

The single-stage NAFNet directly predicts the clean image from a single low-quality observation. As shown in Figure~\ref{fig:discussion-stagewise}, it suppresses the raindrops but retains reflection leakage and produces softened boundaries and local structural distortions around the logo. In contrast, our MAFNet-UR\textsuperscript{3}-L{+} leverages complementary information across multiple observations to reduce residual reflections and preserve the logo boundary more faithfully.

\begin{figure}[t]
  \centering
  \includegraphics[width=\linewidth]{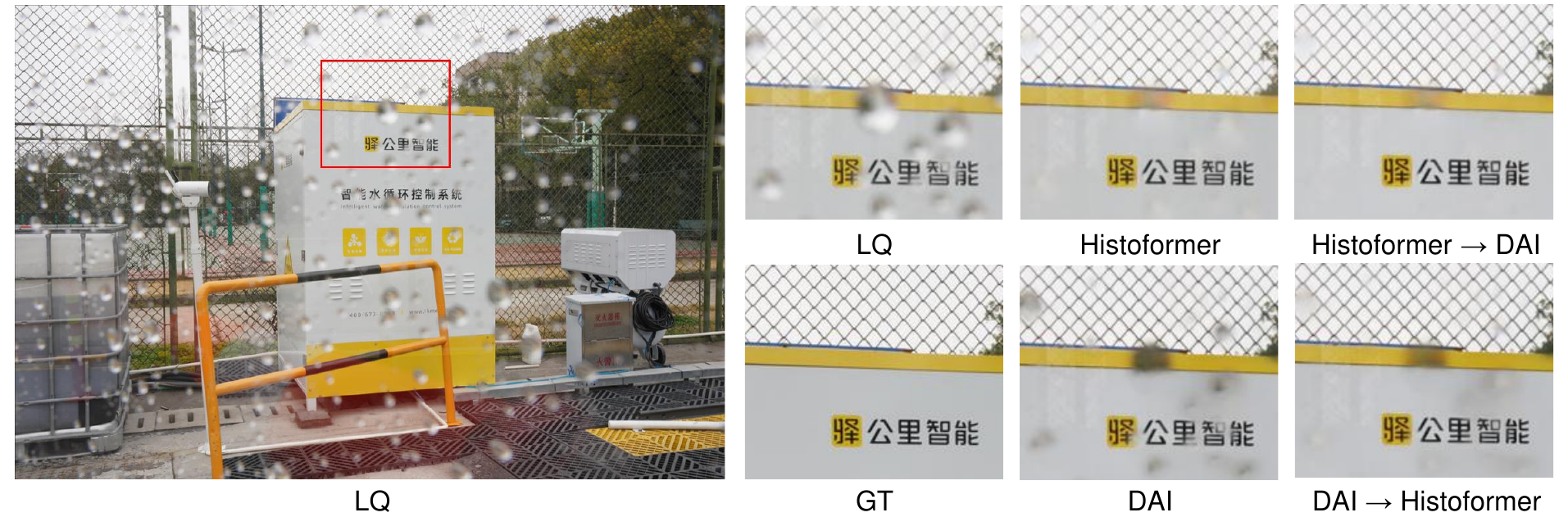}
  \caption{Comparison of raindrop--reflection removal orders.}

  \label{fig:discussion-removal-order}
\end{figure}

\paragraph{Analysis of Removal Orders}
We compare Histoformer~\cite{sun2024restoring} and DAI~\cite{hu2026dereflection} in both cascade orders. Figure~\ref{fig:discussion-removal-order} presents the corresponding qualitative comparison.

The comparison supports raindrop removal before reflection removal for two reasons. First, reflection removal is generally more challenging, as reflection structures overlap semi-transparently with the transmission layer. This tendency is also evident in Figure~\ref{fig:discussion-removal-order}: Histoformer removes most of the raindrops, whereas DAI leaves substantial reflection. We therefore adopt this order, as it is reasonable to remove the easier degradation first. Second, applying DAI first may misinterpret raindrop-induced distortions as reflections, introducing blur and structural errors that Histoformer cannot fully recover. Applying Histoformer first mitigates these distortions and facilitates more reliable transmission–reflection separation by DAI, supporting our raindrop-first, reflection-second cascade.

\section{Conclusion}
\label{sec:conclusion}

In this work, we presented MAFNet-UR\textsuperscript{3}, a three-stage multi-observation framework for scene-consistent raindrop and reflection removal. Stages I and II sequentially remove the two degradations, while Stage III fuses complementary observations using degradation masks and learned spatial confidence before refining a restoration anchor with a magnitude-constrained residual. On RDRF-testing, MAFNet-UR\textsuperscript{3}-L achieves a PSNR of 31.74 dB and a challenge score of 41.1389, outperforming DiffUR\textsuperscript{3} by 2.33 dB and 2.7634, respectively. Our LoViF 2026 UR\textsuperscript{3} Challenge submission ranked third with a final score of 41.1416. Qualitative comparisons and ablation studies further validate the effectiveness of the proposed framework. In particular, they demonstrate consistent performance gains as the number of observations increases. Furthermore, the multi-observation design is naturally extensible to video restoration, where consecutive frames can provide complementary observations of the same scene.

\section*{Acknowledgements}
This work was supported by the GIST ISD program of the Ministry of Science and ICT (KH1160) and the ANCHOR program through the Gwangju ANCHOR Center, funded by the Ministry of Education (MOE) and the Jeonnam--Gwangju Special Metropolitan City, Republic of Korea (2026-ANCHOR-05-001).

\bibliographystyle{splncs04}
\bibliography{main}
\end{document}